\documentclass[final]{cvpr}

\usepackage{times}
\usepackage{epsfig}
\usepackage{graphicx}
\usepackage{amsmath}
\usepackage{amssymb}

\usepackage{xspace}
\usepackage{booktabs}
\usepackage[ruled,vlined]{algorithm2e}
\usepackage{multirow}
\usepackage{xcolor}
\usepackage{caption}
\usepackage{subcaption}

\usepackage[pagebackref=true,breaklinks=true,colorlinks,bookmarks=false]{hyperref}

\def\confYear{CVPR 2021}

\newcommand{\figLabel}{Figure\xspace}

\newcommand{\secLabel}{Section\xspace}
\newcommand{\tblLabel}{Table\xspace}
\newcommand{\mysection}[1]{\vspace{3pt}\noindent\textbf{#1.}}
\newcommand{\supp}{\textbf{supplement}\xspace}

\newcommand{\dense}{$\rhd$\hspace{-2pt}$\lhd$}

\newcommand\blfootnote[1]{%
\begingroup 
\renewcommand\thefootnote{}\footnote{#1}%
\addtocounter{footnote}{-1}%
\endgroup 
}

\begin{document}

\title{PU-GCN: Point Cloud Upsampling using Graph Convolutional Networks} 

\author{Guocheng Qian*
\and
Abdulellah Abualshour* \and
Guohao Li
\and Ali Thabet \and Bernard Ghanem
\\
King Abdullah University of Science and Technology (KAUST)\\
{\tt\small \{guocheng.qian,
abdulellah.abualshour, guohao.li, 
ali.thabet,bernard.ghanem\} @kaust.edu.sa
}
}

\maketitle

\blfootnote{* Guocheng and Abdulellah contributed equally to this work. }
\begin{abstract}
The effectiveness of learning-based point cloud upsampling pipelines heavily relies on the upsampling modules and feature extractors used therein. For the point upsampling module, we propose a novel model called NodeShuffle, which uses a Graph Convolutional Network (GCN) to better encode \emph{local point information} from point neighborhoods. NodeShuffle is \emph{versatile} and can be incorporated into any point cloud upsampling pipeline. Extensive experiments show how NodeShuffle consistently improves state-of-the-art upsampling methods. For feature extraction, we also propose a new \emph{multi-scale point feature extractor}, called Inception DenseGCN. By aggregating features at multiple scales, this feature extractor enables further performance gain in the final upsampled point clouds. We combine Inception DenseGCN with NodeShuffle into a new point upsampling pipeline called PU-GCN. PU-GCN sets new state-of-art performance with much fewer parameters and more efficient inference. Our code is publicly available at \url{https://github.com/guochengqian/PU-GCN}. 
\end{abstract}

\section{Introduction}
\begin{figure}[t]
    \centering
    \includegraphics[page=3,trim = 5mm 40mm 98mm 0mm, clip, width=\columnwidth]{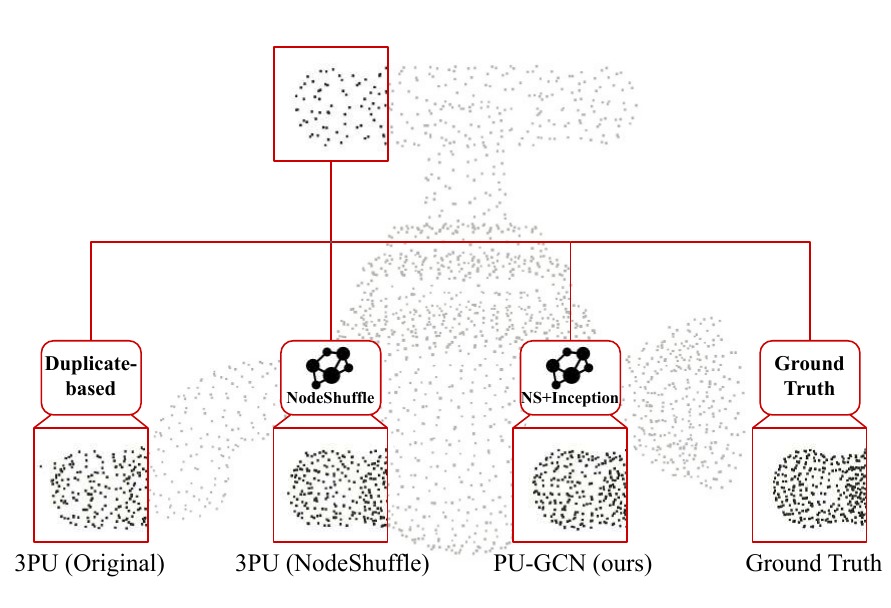}
    \caption{\textbf{Effectiveness of proposed \emph{NodeShuffle} and \emph{Inception DenseGCN}.}
    We propose a Graph Convolutional Network (GCN) based upsampling module \emph{NodeShuffle} and a multi-scale feature extractor \emph{Inception DenseGCN}. Integrating NodeShuffle into the 3PU \cite{Yifan20193pu} upsampling pipeline allows for better upsampling and better structure preservation capability. We propose PU-GCN that combines both Inception DenseGCN and NodeShuffle (NS) upsampling modules. In PU-GCN, Inception DenseGCN can further improve upsampling quality and generate fine-grained details (\eg the neck and ball shape of the faucet). The original 3PU uses duplicate-based upsampling. 
    }
    \label{fig:intro-surface_3pu_up}
\end{figure}

Point clouds are a popular way to represent 3D data. This increasing popularity stems from the increased availability of 3D sensors like LiDAR. Such sensors are now a critical part of important applications in robotics and self-driving cars. However, due to hardware and computational constraints, these 3D sensors often produce sparse and noisy point clouds, which show evident limitations especially for small objects or those far away from the camera. Therefore, point cloud upsampling, the task of converting sparse, incomplete, and noisy point clouds into dense, complete, and clean ones, is attracting much attention. 

Following the success in image super-resolution \cite{Dong2014ImageSU, Lim2017EDSR, Kim2015vdsr, Shi2016RealTimeSI}, deep learning methods now achieve state-of-the-art results in point cloud upsampling
\cite{yu2018ec, yu2018pu, Yifan20193pu, li2019pugan}.  Most deep upsampling pipelines comprise two major components: feature extraction and point upsampling. The performance of the point upsampling component tends to define the effectiveness of the final network. Current methods use either multi-branch MLPs (PU-Net \cite{yu2018pu}) or a duplicate-based approach (3PU \cite{Yifan20193pu} and PU-GAN \cite{li2019pugan}) to upsample 3D points. Multi-branch MLPs operate on each point separately, ignoring any neighborhood information, while duplicate upsampling methods tend to generate point patches similar to the input point clouds. Although the feature extraction modules used in their networks can encode the locality, these shortcomings of the upsampling modules still lead to upsampled point clouds that lack local detail (see \figLabel \ref{fig:intro-surface_3pu_up}). To better represent locality and aggregate the point neighborhood information, we leverage the power of graphs and specifically Graph Convolutional Networks (GCNs). GCNs are considered a powerful tool to process non-Euclidean data, and recent research on point cloud semantic and part segmentation show their power in encoding local and global information \cite{dgcnn, Li2019DeepGCNs, Li2019DeepGCNsMG, thomas2019KPConv}. In this paper, we use GCNs to design a novel and versatile point upsampling module called NodeShuffle (\figLabel \ref{fig:method-upsampling}), which is better equipped at encoding local point information and at learning to generate new points instead of merely replicating parts of the input. 

Point clouds often represent objects of variable part sizes. Using multi-scale features is an effective way to encode this property and is essential for obtaining point clouds of high quality. Recent works like PU-Net \cite{yu2018pu} extract point features at different downsampled levels. While such an architecture can encode multi-scale features, downsampling leads to loss of fine-grained details. In contrast, 3PU \cite{Yifan20193pu} proposes a progressive upsampling network using different numbers of neighbors in subsequent upsampling units. This achieves different receptive fields and encodes multi-scale information. However, 3PU is computationally expensive due to its progressive nature. In this paper, we tackle this multi-scale feature learning problem using a multi-path densely connected GCN architecture called Inception DenseGCN. Following its prevalent usage in image recognition \cite{Szegedy2014inception, Szegedy2015RethinkingTI, Szegedy2016Inceptionv4IA} for the merits of efficient extraction of multi-scale image information, we adopt the Inception architecture to encode multi-scale point features, after it is modified to use densely connected GCNs instead of CNNs. 

\mysection{Contributions}
We summarize our contributions as three-fold. \textbf{(1)} We propose \emph{NodeShuffle}, a novel point cloud upsampling module using graph convolutions. We show how NodeShuffle can be seamlessly integrated into current point upsampling pipelines and consistently improve their performance. \textbf{(2)} We design \emph{Inception DenseGCN}, a feature extraction block that effectively encodes multi-scale information. We combine Inception DenseGCN and NodeShuffle into a new architecture called \emph{PU-GCN}. As compared to the state-of-the-art, PU-GCN achieves better upsampling quality, requires less parameters, and runs faster. Through extensive quantitative and qualitative experiments and for both synthetic and real data, we show the superior performance of PU-GCN. \textbf{(3)} We compile \emph{PU1K}, a new large-scale point cloud upsampling dataset with various levels of shape diversity. We show the challenge of PU1K to current learning-based methods.

\section{Related Work}

\mysection{Learning-based point cloud upsampling} 
Deep learning methods illustrate a promising improvement over their optimization-based counterparts \cite{Alexa2003ComputingAR, Lipman:2007:PPG:1275808.1276405, Huang2013EdgeawarePS, Wu:2015:DPC:2816795.2818073} due to their data-driven nature and the learning capacity of neural networks.
Learning features directly from point clouds was made possible by deep neural networks, such as PointNet \cite{GarciaGarcia2016PointNetA3}, PointNet++ \cite{Qi2017PointNetDH}, DGCNN \cite{dgcnn}, KPConv \cite{thomas2019KPConv}, \etc. 
Yu \etal \cite{yu2018pu} introduced PU-Net, which learns multi-scale features per point and expands the point set via multi-branch MLPs. However, PU-Net needs to downsample the input first to learn multi-scale features, which causes unnecessary resolution loss. 
Yu \etal \cite{yu2018ec} also proposed EC-Net, an edge-aware network for point set consolidation. It uses an edge-aware joint loss to encourage the network to learn to consolidate points for edges. However, EC-Net requires a very expensive edge-notation for training. Wang \etal \cite{Yifan20193pu} proposed 3PU, a progressive network that duplicates the input point patches over multiple steps. 3PU is computationally expensive due to its progressive nature, and it requires more data to supervise the middle stage outputs of the network. Recently, Li \etal \cite{li2019pugan} proposed PU-GAN, a Generative Adverserial Network (GAN) designed to learn upsampled point distributions. While the major contribution and the performance gain is from the discriminator part, the generator architecture receives less attention in their work.
Recently, PUGeo-Net\cite{Qian2020PUGeoNetAG} proposes to upsample points by learning the first and second fundamental forms of the local geometry. However, their method needs additional supervision in the form of normals, which many point clouds like those generated by LIDAR sensors do not come with. Our work tackles upsampling by leveraging a new Inception based module to extract multi-scale information, and by using a novel GCN-based upsampling module to capture local point information. This avoids the need for additional annotations (\eg edges, normals, point clouds at intermediate resolutions, \etc) or a sophisticated discriminator. 

\mysection{3D shape completion}
3D shape completion is intertwined with point cloud upsampling. Researching methods in shape completion can inspire the works of upsampling, and vice versa. Earlier and some recent works on shape completion are based on voxel representations \cite{Wu20153DSA, Dai2017ShapeCU, Han2017HighResolutionSC, Xie2020GRNetGR} and implicit representations \cite{Stutz2018Learning3S}. Point cloud upsampling can also be tackled with these representations, which could be an interesting research direction. More related to point cloud upsampling, point-based completion methods \cite{Yuan2018PCNPC, Yang2018FoldingNetPC, Tchapmi2019TopNetSP, Liu2020MorphingAS, Wang2020SoftPoolNetSD} that directly process on the point clouds (point coordinates and the attributes) have recently shown comparable or even superior performance to those voxel or implicit representation based counterparts.  

\mysection{Graph convolutional networks (GCNs)} 
To cope with the increasing amount of non-Euclidean data in real-world scenarios, a surge of graph convolutional networks \cite{kipf2016semi,hamilton2017inductive,velivckovic2017graph,pham2017column,dgcnn} have been proposed in recent years. Kipf \etal \cite{kipf2016semi} simplify spectral graph convolutions with a first-order approximation. DGCNN \cite{dgcnn} propose EdgeConv to conduct dynamic graph convolution on point clouds. DeepGCNs \cite{Li2019DeepGCNs,Li2019DeepGCNsMG} introduce residual/dense connections and dilated convolutions to GCNs, and successfully trained deep GCN architectures. Previous GCN works mainly investigate basic modules of discriminative models. However, due to the unordered and irregular nature of graph data, generative tasks for this modality remain elusive. Recently, Graph U-Nets \cite{Gao2019GraphU} propose the graph unpooling operation, which can only be used to restore the graph to its original structure and is not able to upsample nodes by an arbitrary ratio. As a focus in this paper, the point upsampling technique, which is an indispensable component for generative models, is under-explored in the GCN domain. We propose a GCN-based upsampling module to tackle the problem. 

\mysection{Multi-scale feature extraction}
Inception architectures \cite{Szegedy2014inception, Szegedy2015RethinkingTI, Szegedy2016Inceptionv4IA} enable superior performance in image recognition at relatively low computational cost. They extract multi-scale information by using different kernel sizes in different paths of the architecture. Inspired by the success of the Inception architecture for CNNs, Kazi \etal\cite{kazi2019inceptiongcn} proposed InceptionGCN, in which feature maps are passed into multiple branches, then each branch applies one graph convolution with a different kernel size. The outputs of these branches are aggregated by concatenation. We also adopt the Inception concept in our work to propose Inception  DenseGCN, a GCN architecture that improves upon InceptionGCN by leveraging dilated graph convolutions, skip connections, and global pooling. 

\section{Methodology}
We propose a novel GCN-based upsampling module (called NodeShuffle) and multi-scale feature extractor (called Inception DenseGCN). We combine Inception DenseGCN and Nodeshuffle into the proposed point cloud upsampling pipeline: PU-GCN. 

\subsection{Upsampling Module: NodeShuffle}
\begin{figure}[ht!]
    \centering
    \includegraphics[page=1,trim = 0mm 0mm 29mm 0mm, clip, width=0.8\columnwidth]{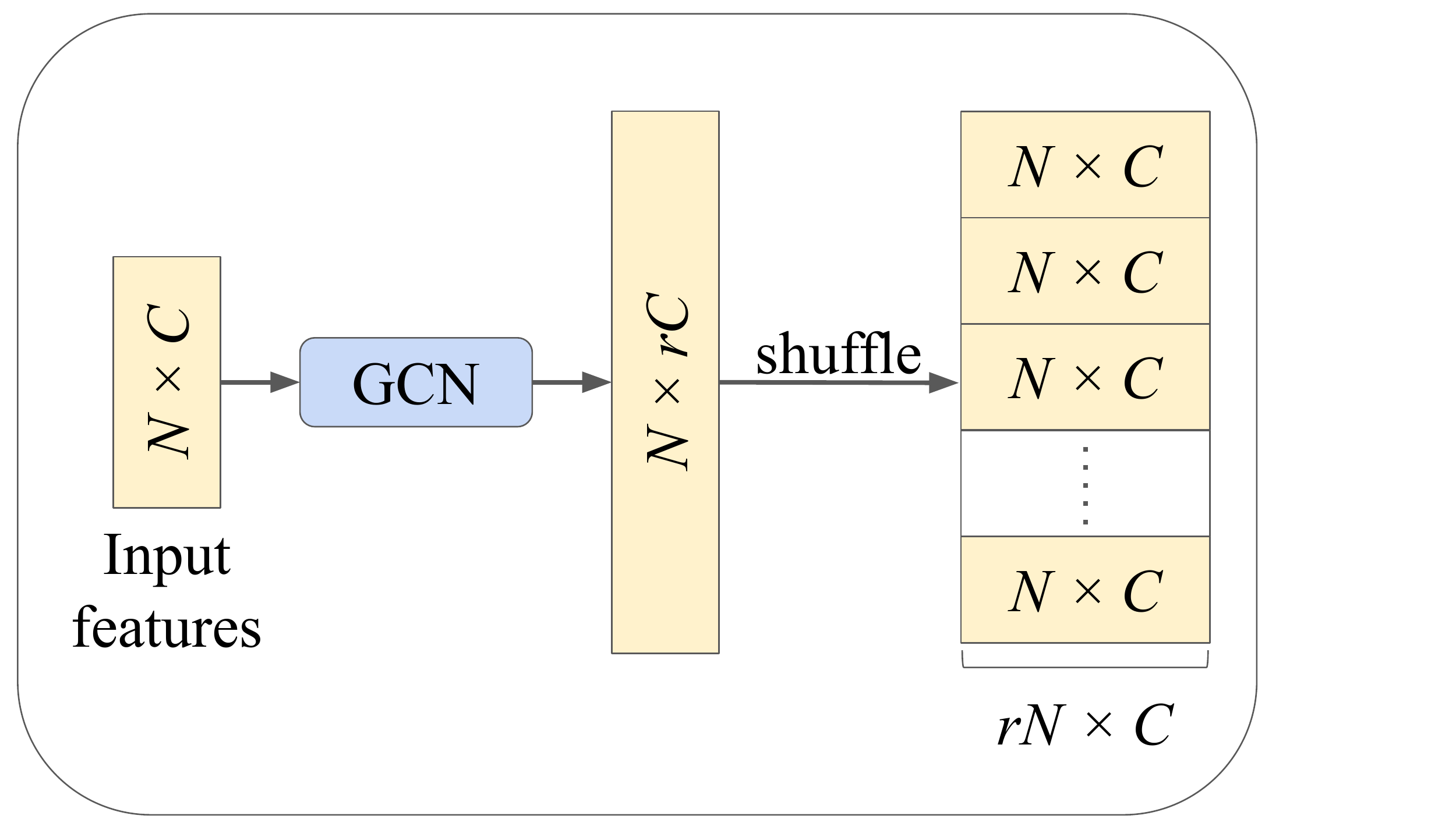}
   \caption{\textbf{Upsampling module: NodeShuffle.} We expand the number of input features by $r$ times using a GCN layer, and then apply a shuffle operation to rearrange the feature map. $r$ denotes the upsampling ratio.
   }
\label{fig:method-upsampling}
\end{figure}

Inspired by PixelShuffle \cite{Shi2016RealTimeSI} from the image super-resolution literature, we propose \emph{NodeShuffle} to effectively upsample point clouds. NodeShuffle is a graph convolutional upsampling layer, illustrated in \figLabel \ref{fig:method-upsampling}.
Given node features $\mathcal{V}_l$ with shape $N \times C$, the NodeShuffle operator will output the new node features $\mathcal{V}_{l+1}$ with shape $rN \times C$ as follows ($r$ is the upsampling ratio):
\begin{equation}
   \mathcal{V}_{l+1}= \mathcal{PS}(\mathcal{W}_{l+1} * \mathcal{V}_l +b_{l+1}),
\end{equation}
where $\mathcal{PS}$ is a periodic shuffling operator that rearranges the graph (\eg point features) of shape $N \times rC$ to $rN \times C$. 
The NodeShuffle operation can be divided into two steps. (1) Channel expansion: we use a 1 layer GCN to expand node features $\mathcal{V}_l$ to shape $N \times rC$ using learnable parameters $\mathcal{W}_{l+1}$ and $b_{l+1}$. (2) Periodic shuffling: we rearrange the output of channel expansion to shape $rN \times C$.

In contrast to multi-branch MLPs \cite{yu2018pu} or duplicate-based upsampling \cite{Yifan20193pu,li2019pugan}, NodeShuffle leverages graph convolutions instead of CNNs. Although GCNs are common modules for feature extraction, to the best of our knowledge, we are the first to introduce a GCN-based upsampling module. Our GCN design choice stems from the fact that GCNs enable our upsampler to encode spatial information from point neighborhoods and learn new points from the latent space rather than simply duplicating the original points (as done in 3PU \cite{Yifan20193pu} and PU-GAN \cite{li2019pugan}) or copying points after different transforms through multi-branch MLPs (as done in PU-Net\cite{yu2018pu}). A more detailed comparison of these upsampling modules can be found in the \supp. 

\subsection{Multi-scale Features: Inception DenseGCN}
Point clouds scanned using 3D sensors often include objects of various sizes and point resolutions. In order to encode the multi-scale nature of point clouds, we propose a new \emph{Inception DenseGCN} feature extractor, which effectively integrates the densely connected GCN (DenseGCN) module from DeepGCNs \cite{Li2019DeepGCNs} into the Inception module from GoogLeNet \cite{Szegedy2014inception}. 
Residual and dense connections have proven to be useful at increasing point cloud processing performance \cite{Li2019DeepGCNs}.
We favor dense over residual connections here, since the former utilizes features from previous layers, as well as different inception paths.

\begin{figure}[t]
\centering
    {
    \includegraphics[page=2,trim =0mm 0mm 80mm 0mm, clip, width=0.6\columnwidth]{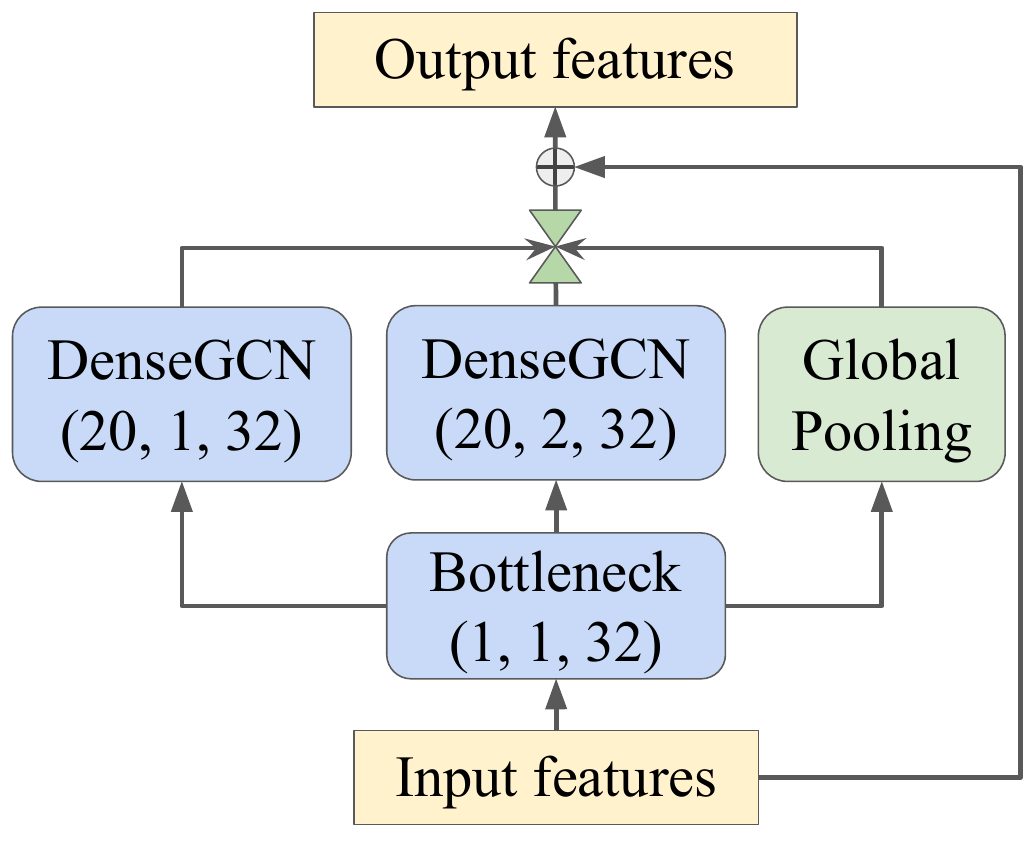}}
    \caption{\textbf{Multi-scale feature extractor: Inception DenseGCN.} We use the parameters ($k$, $d$, $c$) to define a DenseGCN block. $k$ is the number of neighbors (kernel size), $d$ is the dilation rate, and $c$ is the number of output channels. Green \dense ~denotes feature-wise concatenation. Note that the $d$ is different in the different DenseGCN blocks to achieve the goal of multi-scale feature extraction. 
}
    \label{fig:method-inception_densegcn}
\end{figure}

\figLabel \ref{fig:method-inception_densegcn} illustrates our Inception DenseGCN block. 
The input features are compressed by a bottleneck layer (single layer MLP) at first to reduce the computation in subsequent layers. 
The compressed features are passed into two parallel DenseGCN blocks. Each DenseGCN block is composed of three layers of densely connected dilated graph convolutions (introduced in DeepGCNs \cite{Li2019DeepGCNs}). DenseGCN is defined by the number of node neighbors $k$ (kernel size for GCNs), dilation rate $d$, and growth rate $c$. 
The two DenseGCN blocks inside the Inception DenseGCN have the same kernel size (20) but different dilation rate (1, and 2, respectively) to gain different respective fields without increasing the kernel sizes and FLOPs.
Similar to dilated convolutions in the 2D case, the dilated graph convolution efficiently increases the receptive field using the same kernel size without reducing spatial resolution.
Additionally, we add a global pooling layer to extract global contextual information. The DenseGCN blocks and the global pooling layer target different receptive fields and therefore allow the Inception module to extract multi-scale information. Each Inception block outputs a concatenation of the DenseGCN blocks and the global pooling layer in addition to the input features. KNN is used to build the graph, whose nodes are the points and edges define the $K$ nearest point neighborhood. Note that KNN is only computed once at the first layer of the Inception DenseGCN block.
Experiments show that making the graph dynamic in the later layers inside an Inception DenseGCN module does not actually impact performance much, but it does increase the model's computational footprint. So, the graph structure is designed to be shared in subsequent DenseGCN layers inside Inception DenseGCN.

The InceptionGCN proposed by Kazi \etal\cite{kazi2019inceptiongcn} is as simple as concatenating multiple GCN layers with different kernel sizes. In contrast, our Inception DenseGCN leverages DenseGCN, global pooling, and skip connections to improve the performance. We further use bottleneck layers and dilated graph convolutions to reduce the computational burden.

\subsection{PU-GCN Architecture}
\begin{figure*}[ht!]
    \centering
    \includegraphics[page=3,trim = 0mm 95mm 0mm 0mm, clip, width=\textwidth]{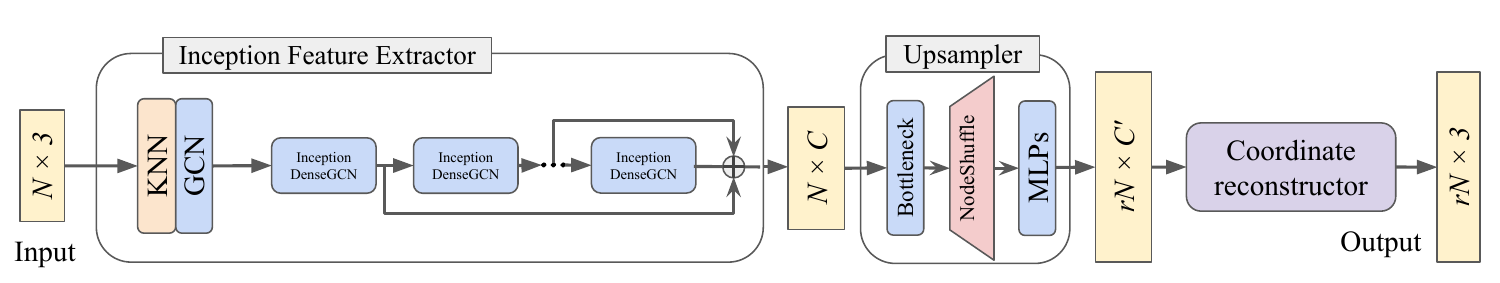}
   \caption{
    \textbf{PU-GCN architecture.}  PU-GCN uses an inception feature extractor consisting of one or more  Inception DenseGCN blocks, followed by the NodeShuffle based upsampler, and a coordinate reconstructor. 
   }
\label{fig:method-arch}
\end{figure*}

We combine Inception DenseGCN, NodeShuffle, and a standard coordinate reconstructor into a new upsampling pipeline called PU-GCN (\figLabel \ref{fig:method-arch}). Given a point cloud of size $N \times 3$, PU-GCN computes point features of size $N \times C$ using the Inception DenseGCN feature extractor. Then, the upsampler transforms the $N \times C$ features to $rN \times C^{\prime}$ using NodeShuffle. Finally, the coordinate reconstructor generates the $rN \times 3$ upsampled point cloud. We use EdgeConv (proposed in DGCNN \cite{dgcnn}) as the default GCN layer. 

\mysection{Inception feature extractor} 
We use 1 GCN layer at the beginning of PU-GCN to embed the 3D coordinates into latent space. 
The point embeddings are passed through several Inception DenseGCN blocks. We use two such blocks by default in PU-GCN, and we experiment with the number of Inception DenseGCN blocks in \secLabel \ref{sec:ablation}.
The outputs of these blocks are combined with residual connections. The final output is passed to our NodeShuffle upsampler. 

\mysection{Upsampler}
Our upsampler consists of three components: a bottleneck layer, an upsampling module, and a feature compression layer. Given the input features, we first shrink the input to $N \times C$ by a bottleneck layer (MLP) so as to reduce the computation. Then, we use the proposed NodeShuffle to generate denser features of size $rN \times C$. Finally, we use two sets of MLPs to compress the features to $rN \times C'$.  

\mysection{Coordinate reconstructor} 
We reconstruct points from latent space to coordinate space, resulting in the desired denser point cloud of size $rN \times 3$. We use the same coordinate reconstruction approach as PU-GAN \cite{li2019pugan}, in which 3D coordinates are regressed using two sets of MLPs. 

\section{Experiments}
We compile a large-scale dataset for point cloud upsampling called PU1K. Quantitative and qualitative results on PU-GAN\cite{li2019pugan}'s dataset as well as PU1K show the superior performance of PU-GCN. We also conduct extensive ablation studies to show the benefits of the  proposed Inception DenseGCN and NodeShuffle upsampling modules.  

\subsection{Datasets}
We propose a new dataset for point cloud upsampling, denoted as \emph{PU1K}. 
It is nearly 8 times larger than the largest publicly available dataset collected by PU-GAN \cite{li2019pugan}. PU1K consists of 1,147 3D models split into 1020 training samples and 127 testing samples. The training set contains 120 3D models compiled from PU-GAN's dataset \cite{li2019pugan}, in addition to 900 different models collected from ShapeNetCore \cite{shapenet}. The testing set contains 27 models from PU-GAN and 100 more models from ShapeNetCore. The models from ShapeNetCore are chosen from 50 different categories. We randomly choose 200 models from each category to obtain 1,000 different models with various shape complexities to encourage diversity. Overall, PU1K covers a large semantic range of 3D objects and includes simple, as well as complex shapes. To show the value of our proposed dataset, we compare our PU-GCN with previous approaches on both PU1K and the latest dataset proposed by PU-GAN, which contains only 120 3D models for training and 27 models for testing. 

For training and testing and following common practice, we use Poisson disk sampling from the original meshes to generate pairs of input and ground truth point clouds. For training, we crop 50 patches from each 3D model as the input to the network. In total, we obtain 51,000 training patches in PU1K. Each patch consists of 256 points as low resolution input and 1024 points as ground truth. As for testing data, we generate pairs of input point cloud (2048 points) and ground truth (8096 points). In testing, we use farthest point sampling at first to sample overlapping patches (patch size is 256) of the input point cloud and ensure coverage of the entire input. The final result is obtained by first merging the overlapping patch outputs and then resampling with farthest point sampling. We will open-source the original meshes of PU1K and the sampled point clouds to standardize the dataset. More details of PU1K and a comparison with PU-GAN's dataset are in \supp.

\subsection{Loss Function and Evaluation Metrics}
\mysection{Loss Function}
We use the Chamfer distance loss to minimize the distance of the predicted point cloud and the referenced ground truth in our experiments:
\begin{equation} \label{chamfer_loss}
     \footnotesize
     \begin{split}
     C(P, Q) = \frac{1}{|P|} \sum_{p \in P} \min_{q \in Q} ||p-q||^{2}_{2} + \frac{1}{|Q|} \sum_{q \in Q} \min_{p \in P} ||p-q||^{2}_{2}   
     \end{split}
\end{equation}
where $P$ is the predicted point cloud, $Q$ is the ground truth, $p$ is a 3D point from $P$, and $q$ is a 3D point from $Q$. The operator $||\cdot||^{2}_{2}$ denotes the squared Euclidean norm. 

\mysection{Evaluation Metrics}
Following previous work, we use the Chamfer distance (CD), Hausdorff distance (HD), and point-to-surface distance (P2F) w.r.t ground truth meshes as evaluation metrics. The smaller the metrics, the better the performance. We also report the parameter size (Params.) and the inference time. The inference time is reported as the average inference time (over the whole test set in 5 runs) for a model processing one patch containing 256 points. All models are tested on the same computer with one NVIDIA TITAN 2080Ti GPU and an Intel Xeon E5-2680 CPU. 

\subsection{Implementation Details} \label{subsec:implementation}
We train PU-GCN for 100 epochs with batch size 64 on an NVIDIA TITAN 2080Ti in all the experiments. We optimize using Adam with a learning rate of 0.001 and beta 0.9. Similar to previous work, we perform point cloud normalization and augmentation (rotation, scaling, and random perturbations).  We train PU-Net \cite{yu2018pu}, 3PU \cite{Yifan20193pu}, and our PU-GCN on both the PU1K dataset and PU-GAN's dataset. Here, we note that we are unable to reproduce PU-GAN's \cite{li2019pugan} results from the code made available by its authors, most probably because of the unstable nature of the inherent GAN architecture in PU-GAN. Therefore, we only compare with PU-GAN on PU-GAN's dataset using their provided pre-trained model. The very recent work PUGeo-Net\cite{Qian2020PUGeoNetAG} is not included in the comparison for the following reasons: (1) their code and pre-trained model are not available; (2) they only conduct experiments on their own dataset that has not been released; (3) they need additional supervision in the form of normals, which are not directly accessible in point clouds. As suggested by previous methods, we use the model from the last epoch to perform the evaluation. We report results using a $\times 4$ upsampling rate, \ie $r=4$. We did not experiment with a large upsampling factor like $r=16$, but one can simply apply our pretrain models twice. 

\begin{table}[t]
\caption{\textbf{Comparison of PU-GCN vs. state-of-the-art on PU-GAN's dataset.} 
PU-GCN outperforms PU-Net, 3PU, and PU-GAN on nearly all metrics with the least parameters and fastest inference. \textbf{Bold} denotes the best performance.
}
\vspace{-10pt}
\label{tab:sota_comparasion_results}
\centering
\resizebox{\columnwidth}{!}{%
\begin{tabular}{l|ccccc}
\toprule
\multirow{2}{*}{\textbf{Network}}
& \multicolumn{1}{c}{\textbf{CD}$ \downarrow$} & \multicolumn{1}{c}{\textbf{HD}$ \downarrow$} & \multicolumn{1}{c}{\textbf{P2F}$ \downarrow$} & 
\multicolumn{1}{c}{\textbf{Param.}} & 
\multicolumn{1}{c}{\textbf{Time}} \\
& \multicolumn{1}{c}{$10^{-3}$} & \multicolumn{1}{c}{$10^{-3}$} & \multicolumn{1}{c}{$10^{-3}$} &
\multicolumn{1}{c}{Kb} & \multicolumn{1}{c}{ms} \\
\midrule
PU-Net \cite{yu2018pu} & 
0.556	& 4.750 &   4.678 & 814.3 & 10.04 \\ 
3PU \cite{Yifan20193pu}  &
0.298	& 4.700 & 2.855 & 76.2	& 10.86 \\
PU-GAN \cite{li2019pugan}   &  
0.280 &	4.640 &	\textbf{2.330} & 684.2 & 14.28 \\
\midrule
\textbf{PU-GCN} & 
\textbf{0.258}	&\textbf{1.885}	& 2.721 & \textbf{76.0} & \textbf{8.83}\\
\bottomrule
\end{tabular}
}
\vspace{-10pt}
\end{table}

\begin{figure*}[h!]
\begin{center}
    \begin{subfigure}{.155\textwidth}
        \centering
        \includegraphics[page=1,trim = 30mm 70mm 20mm 15mm, width=1.0\textwidth]{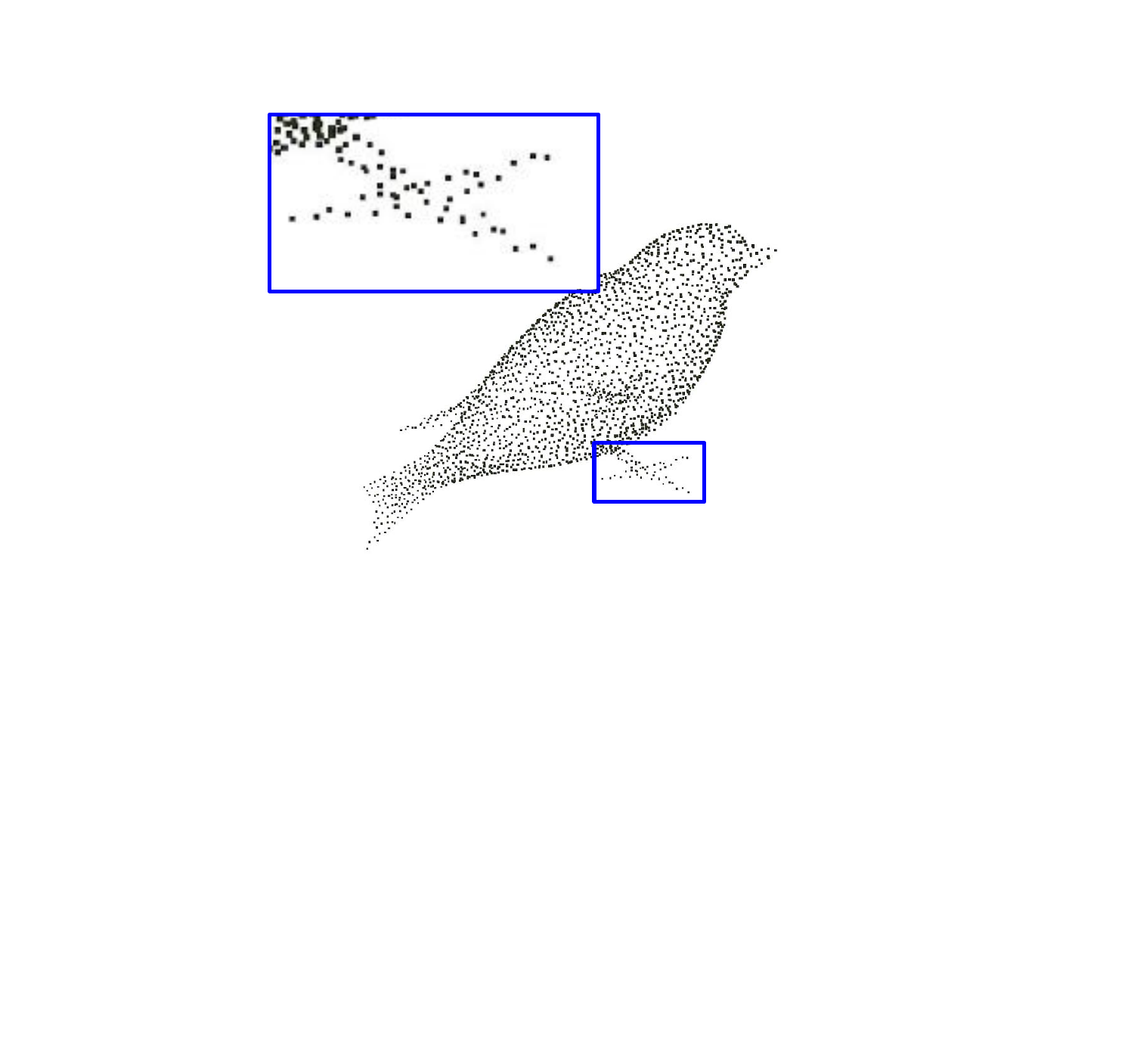}
    \end{subfigure}
    \begin{subfigure}{.155\textwidth}
        \centering
        \includegraphics[page=2,trim = 30mm 70mm 20mm 15mm,
        clip, width=1.0\textwidth]{figures/pugcn_cvpr21_qualitative_reduced.pdf}
    \end{subfigure}
    \begin{subfigure}{.155\textwidth}
        \centering
        \includegraphics[page=3,trim = 30mm 70mm 20mm 15mm,
        clip, width=1.0\textwidth]{figures/pugcn_cvpr21_qualitative_reduced.pdf}
    \end{subfigure}
    \begin{subfigure}{.155\textwidth}
        \centering
        \includegraphics[page=4,trim = 30mm 70mm 20mm 15mm, width=1.0\textwidth]{figures/pugcn_cvpr21_qualitative_reduced.pdf}
    \end{subfigure}
    \begin{subfigure}{.155\textwidth}
        \centering
        \includegraphics[page=5,trim = 30mm 70mm 20mm 15mm, width=1.0\textwidth]{figures/pugcn_cvpr21_qualitative_reduced.pdf}
    \end{subfigure}
    \begin{subfigure}{.155\textwidth}
        \centering
        \includegraphics[page=6,trim = 30mm 70mm 20mm 15mm, width=1.0\textwidth]{figures/pugcn_cvpr21_qualitative_reduced.pdf}
    \end{subfigure}

    \begin{subfigure}{.155\textwidth}
        \centering
        \includegraphics[page=7,trim = 60mm 55mm 30mm 10mm,, clip, width=1.0\textwidth]{figures/pugcn_cvpr21_qualitative_reduced.pdf}
    \end{subfigure}
    \begin{subfigure}{.155\textwidth}
        \centering
        \includegraphics[page=8,trim = 60mm 55mm 30mm 10mm, clip, width=1.0\textwidth]{figures/pugcn_cvpr21_qualitative_reduced.pdf}
    \end{subfigure}
    \begin{subfigure}{.155\textwidth}
        \centering
        \includegraphics[page=9,trim = 60mm 55mm 30mm 10mm, clip, width=1.0\textwidth]{figures/pugcn_cvpr21_qualitative_reduced.pdf}
    \end{subfigure}
    \begin{subfigure}{.155\textwidth}
        \centering
        \includegraphics[page=10,trim = 60mm 55mm 30mm 10mm, clip, width=1.0\textwidth]{figures/pugcn_cvpr21_qualitative_reduced.pdf}
    \end{subfigure}
    \begin{subfigure}{.155\textwidth}
        \centering
        \includegraphics[page=11,trim = 60mm 55mm 30mm 10mm, clip, width=1.0\textwidth]{figures/pugcn_cvpr21_qualitative_reduced.pdf}
    \end{subfigure}
    \begin{subfigure}{.155\textwidth}
        \centering
        \includegraphics[page=12,trim = 60mm 55mm 30mm 10mm, clip, width=1.0\textwidth]{figures/pugcn_cvpr21_qualitative_reduced.pdf}
    \end{subfigure}

    \begin{subfigure}{.155\textwidth}
        \centering
        \includegraphics[page=13,trim = 35mm 40mm 30mm 10mm, clip, width=1.0\textwidth]{figures/pugcn_cvpr21_qualitative_reduced.pdf}
        \caption{Input}
        \label{fig:exp-qual_results-input}
    \end{subfigure}
    \begin{subfigure}{.155\textwidth}
        \centering
        \includegraphics[page=14,trim = 35mm 40mm 30mm 10mm, clip, width=1.0\textwidth]{figures/pugcn_cvpr21_qualitative_reduced.pdf}
        \caption{PU-Net \cite{yu2018pu}}
        \label{fig:exp-qual_results-punet}
    \end{subfigure}
    \begin{subfigure}{.155\textwidth}
        \centering
      \includegraphics[page=15,trim = 35mm 40mm 30mm 10mm, clip, width=1.0\textwidth]{figures/pugcn_cvpr21_qualitative_reduced.pdf}
        \caption{3PU \cite{Yifan20193pu}}
        \label{fig:exp-qual_results-3pu}
    \end{subfigure}
    \begin{subfigure}{.155\textwidth}
        \centering
        \includegraphics[page=16,trim = 35mm 40mm 30mm 10mm, clip, width=1.0\textwidth]{figures/pugcn_cvpr21_qualitative_reduced.pdf}
        \caption{PU-GAN \cite{li2019pugan}}
        \label{fig:exp-qual_results-pugan}
    \end{subfigure}
    \begin{subfigure}{.155\textwidth}
        \centering
        \includegraphics[page=17,trim = 35mm 40mm 30mm 10mm, clip, width=1.0\textwidth]{figures/pugcn_cvpr21_qualitative_reduced.pdf}
         \caption{PU-GCN (Ours)}
        \label{fig:exp-qual_results-ours}
    \end{subfigure}
    \begin{subfigure}{.155\textwidth}
        \centering
        \includegraphics[page=18,trim = 35mm 40mm 30mm 10mm, clip, width=1.0\textwidth]{figures/pugcn_cvpr21_qualitative_reduced.pdf}
         \caption{GT}
        \label{fig:exp-qual_results-gt}
    \end{subfigure}
\end{center}
\vspace{-15pt}
\caption{\textbf{Qualitative upsampling results.} We show the $\times 4$ upsampled results of input point clouds (2048 points) when processed by different upsampling methods (PU-Net, 3PU, PU-GAN and our PU-GCN). PU-GCN produces the best results overall, while generating less outliers and preserving fine-grained local details (refer to close-ups). }
\label{fig:exp-qual_results}
\end{figure*}

\subsection{Quantitative and Qualitative Results}
\mysection{Quantitative results on PU-GAN's dataset} 
\tblLabel \ref{tab:sota_comparasion_results} reports the performance of our PU-GCN compared to PU-Net\cite{yu2018pu}, 3PU\cite{Yifan20193pu}, and PU-GAN\cite{li2019pugan} on PU-GAN's dataset. PU-GCN maintains significant improvement over 3PU and PU-Net in all metrics, showing the importance of the Inception DenseGCN feature extractor and the NodeShuffle upsampling module. Although PU-GAN leverages an adversarial loss for performance gains, we also outperform PU-GAN in terms of CD and HD, without the need for a GAN architecture. It is also important to mention that PU-GCN is the most parameter-saving and the most efficient architecture among all the models. Compared to PU-GAN, our PU-GCN  uses only about 10\% of the parameters and speeds up inference time by more than 40\%. For the P2F metric, we do not outperform PU-GAN quantitatively. However, the qualitative results in \figLabel \ref{fig:exp-qual_results} show that PU-GCN generates fewer outliers and higher quality local and fine-grained details (\eg~the legs of the bird in row 1) than PU-GAN in general, even though these upsampled point clouds have higher P2F values than those of PU-GAN. This discrepancy indicates that P2F might not be as reliable a metric as CD and HD for point cloud upsampling, especially since the latter distances are meant for point clouds (the natural form of the input/output in training) and not meshes (as is the case for P2F). More examples and an analysis of this discrepancy can be found in the \supp.

\mysection{Quantitative results on the PU1K dataset}
\tblLabel \ref{tab:sota_comparasion_PU1K} compares PU-GCN against PU-Net and 3PU on PU1K. We do not compare against PU-GAN on this dataset for the reason mentioned in Section \ref{subsec:implementation}. We observe that all methods achieve lower performance overall (especially in CD and HD) when trained and evaluated on PU1K as compared to PU-GAN's dataset. As such, PU1K presents a challenge to state-of-the-art methods compared to the much smaller and less diverse PU-GAN's dataset. PU-GCN also clearly outperforms PU-Net and 3PU in all three metrics on this new challenging dataset. 

\begin{table}[t]
\caption{\textbf{Comparison of PU-GCN vs. state-of-the-art on PU1K.} 
PU-GCN outperforms PU-Net and 3PU. We do not compare against PU-GAN on this dataset, since we are unable to reproduce PU-GAN results from the publicly available code. \textbf{Bold} denotes the best performance.
}
\vspace{-10pt}
\label{tab:sota_comparasion_PU1K}
\centering
\resizebox{0.9\columnwidth}{!}{%
\begin{tabular}{l|ccccc}
\toprule
\multirow{2}{*}{\textbf{Network}}
& \multicolumn{1}{c}{\textbf{CD}$\downarrow$} & \multicolumn{1}{c}{\textbf{HD}$\downarrow$} & \multicolumn{1}{c}{\textbf{P2F}$\downarrow$} & 
\multicolumn{1}{c}{\textbf{Param.}} & 
\multicolumn{1}{c}{\textbf{Time}} \\
&\multicolumn{1}{c}{$10^{-3}$} & \multicolumn{1}{c}{$10^{-3}$} & \multicolumn{1}{c}{$10^{-3}$} &
\multicolumn{1}{c}{Kb} & \multicolumn{1}{c}{ms} \\
\midrule
PU-Net\cite{yu2018pu} & 1.155          & 15.170         & 4.834          & 814.3         & 10.04         \\
3PU\cite{Yifan20193pu}    & 0.935          & 13.327         & 3.551          & 76.2          & 10.86         \\
\midrule
\textbf{PU-GCN}                        & \textbf{0.585} & \textbf{7.577} & \textbf{2.499} & \textbf{76.0} & \textbf{8.83}\\
\bottomrule
\end{tabular}
}
\vspace{-10pt}
\end{table}

\mysection{Qualitative results} 
\figLabel \ref{fig:exp-qual_results} shows qualitative upsampling results generated by PU-GCN and the state-of-the-art methods. We note that all models used here are trained on PU-GAN's dataset for a fair comparison with PU-GAN.  Upsampled point clouds and their close-ups show that PU-GCN produces fewer outliers, while preserving more fine-grained details. Specifically, close-ups of the bird point cloud (top row) show that PU-GCN successfully upsamples intricate structures of input points. In addition, the statue (second row) shows how PU-GCN succeeds in upsampling with much fewer outliers. We also observe that other methods tend to merge originally separate structures as shown in the example of the clock (third row), while our method preserves this separation with high quality. The qualitative results clearly show the effectiveness of our multi-scale feature extractor (Inception DenseGCN) and our GCN based upsampling module (NodeShuffle) in capturing detailed local information. More qualitative results (especially for models trained on PU1K dataset) and a discussion of some failure cases can be found in the \supp. 

\subsection{Upsampling Real-Scanned Point Clouds}
\begin{figure*}[htb!]
\begin{center}
    \begin{subfigure}{.24\textwidth}
        \centering
        \includegraphics[page=1,trim = 0mm 45mm 0mm 0mm, clip, width=1.0\textwidth]{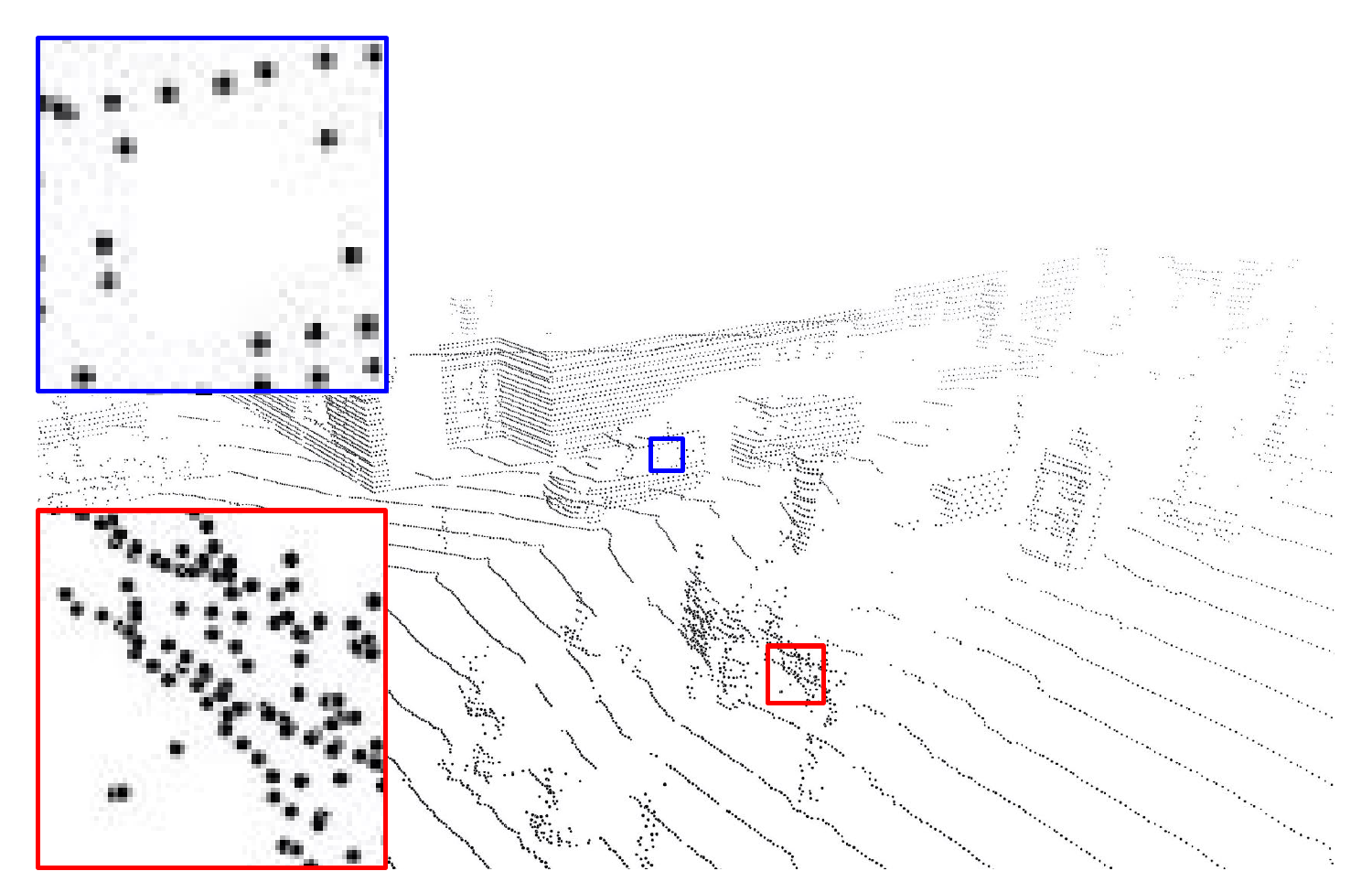}
        \caption{Input}
    \end{subfigure}
    \begin{subfigure}{.24\textwidth}
        \centering
        \includegraphics[page=2,trim = 0mm  45mm 0mm 0mm, clip, width=1.0\textwidth]{figures/pugcn_cvpr21_kitti_reduced.pdf}
        \caption{3PU \cite{Yifan20193pu}}
    \end{subfigure}
    \begin{subfigure}{.24\textwidth}
        \centering
       \includegraphics[page=3,trim = 0mm  45mm 0mm 0mm, clip, width=1.0\textwidth]{figures/pugcn_cvpr21_kitti_reduced.pdf}
         \caption{PU-GAN \cite{li2019pugan}}
    \end{subfigure}
    \begin{subfigure}{.24\textwidth}
        \centering
       \includegraphics[page=4,trim = 0mm  45mm 0mm 0mm, clip, width=1.0\textwidth]{figures/pugcn_cvpr21_kitti_reduced.pdf}
     \caption{PU-GCN (ours)}
    \end{subfigure}

\vspace{-5pt}    
\end{center}
\vspace{-15pt}
\caption{\textbf{Upsampling real-scanned point clouds from KITTI \cite{Geiger2012CVPR}}. PU-GCN preserves intricate structures and generates fine-grained details (\eg the window of the car, the pedal of the motorcycle, \etc). Please zoom in to see details.}
\label{fig:exp-kitti}
\vspace{-10pt}
\end{figure*}

\figLabel \ref{fig:exp-kitti} compares PU-GCN against its two most competitive upsampling methods (3PU and  PU-GAN) on real-scanned data from the KITTI dataset \cite{Geiger2012CVPR}. KITTI contains challenging real-world scenes and objects.
All the models are trained on PU-GAN's dataset for a fair comparison with PU-GAN. To ease visualization and since it consistently produces worse results, PU-Net \cite{yu2018pu} is not included here. We observe that the other methods tend to produce more outliers and overfill holes (\eg the first close up on the window of the car), while PU-GCN encodes local information well and preserves these fine-grained details. Our method also does very well when upsampling objects of interest, such as motorcycle as shown in the second close up. While other methods tend to merge the pedal and body of the motorcycle and ruin its original shape, PU-GCN successfully produces a higher level of detail and structure. More examples of real-scanned point clouds from different datasets can be found in the \supp. 

\subsection{Ablation Study}
\label{sec:ablation}
We conduct ablation studies on NodeShuffle and Inception DenseGCN. The baseline model is PU-GCN equipped with two Inception DenseGCN and the NodeShuffle upsampler. All the models are trained and evaluated on PU1K. 

\mysection{Inception Modules}
We validate the effectiveness of our Inception DenseGCN by replacing it with a DenseGCN. The results in \tblLabel \ref{tbl:ablation_inception} show that our Inception DenseGCN outperforms the DenseGCN in all metrics. The second row demonstrates that the Inception DenseGCN with multiple receptive fields achieved by using two DenseGCN blocks improves the upsampling quality a lot with similar latency, compared to Inception DenseGCN with only a single DenseGCN block. Additionally, using DenseGCN in Inception DenseGCN works better by a large margin than using GCN (Incpetion GCN) as expected (third row).  The fourth row shows that using dilated graph convolutions can achieve better performance, while being more efficient compared to the Inception module using different kernel sizes. We further show that using residual connections and global pooling inside Inception DenseGCN improves performance with a negligible effect on inference speed (row 5 and row 6). We also study the effect of the number of Inception DenseGCN blocks. PU-GCN with two Inception blocks outperforms PU-GCN with only one Inception block, thus showing that inserting one more Inception DenseGCN block can further improve upsampling performance. We also experimented with using more blocks, but performance did not improve while computational complexity did. Therefore, we use two Inception DenseGCN blocks in PU-GCN by default. 

\begin{table}[htb!]
\centering
\caption{\textbf{Ablation study on Inception DenseGCN}. Inception DenseGCN performs better than a DenseGCN. Using two DenseGCN blocks in Inception DenseGCN  to extract multi-scale information is better than using only one block. As expected, using DenseGCN in Inception DenseGCN works better than using GCN (Inception GCN). Using dilated graph convolution instead of regular graph convolution with different kernel sizes also achieves better performance with faster inference. Residual connections and global pooling inside Inception DenseGCN further improve performance. Increasing the number of Inception DenseGCN blocks tends to improve PU-GCN performance.
}
\vspace{-10pt}
\resizebox{\columnwidth}{!}{%
\setlength{\tabcolsep}{5pt}
\begin{tabular}{l|ccccc}
\toprule
\multirow{2}{*}{\textbf{Ablation}} & \multicolumn{1}{c}{\textbf{CD}$\downarrow$} & 
\multicolumn{1}{c}{\textbf{HD}$\downarrow$} & 
\multicolumn{1}{c}{\textbf{P2F}$\downarrow$} &
\multicolumn{1}{c}{\textbf{Param.}} & 
\multicolumn{1}{c}{\textbf{Time}} 
\\ 
& \multicolumn{1}{c}{\textbf{$10^{-3}$}} & \multicolumn{1}{c}{\textbf{$10^{-3}$}} & \multicolumn{1}{c}{\textbf{$10^{-3}$}}
& \multicolumn{1}{c}{Kb}
& \multicolumn{1}{c}{ms}
\\
\midrule
DenseGCN (w/o Inception) & 0.753 & 10.691  & 3.103  & \textbf{56.16} & 9.89         \\
single DenseGCN block & 0.630 & 9.428 & 2.608 & 56.32 & 8.82 \\
Inception GCN & 0.675 & 9.951  & 2.723  & 59.59 & 8.65         \\
\midrule
w/o dilated graph convolution & 0.624 & 8.871 & 2.530 & 75.97 &  8.91\\
\midrule
w/o residual connection & 0.621 & 8.979 & 2.603 & 75.97 & 8.74  \\
w/o global pooling & 0.663 & 9.875 & 2.665  & 75.88  & 8.74  \\
\midrule
1 Inception DenseGCN & 0.639 & 9.051 & 2.582  & 59.33 & \textbf{6.40} \\
PU-GCN & \textbf{0.585} & \textbf{7.577} & \textbf{2.499} & 75.97 & 8.83 \\    
\bottomrule
\end{tabular}
}
\label{tbl:ablation_inception}
\end{table}

\mysection{Upsampling Modules}
We show the effectiveness of our NodeShuffle by integrating it into different upsampling architectures and replacing the original upsampling modules. Results in \tblLabel \ref{tab:up_ablation_results} clearly show  that NodeShuffle helps both PU-Net and 3PU reach better performance with less parameters and negligible computational overhead ($\leq$ 1ms latency).
We also study the effectiveness of GCN inside NodeShuffle by replacing it with a set of MLPs, called MLPShuffle. GCN outperforms the MLPs counterpart. 
\tblLabel \ref{tab:up_ablation_results} also shows that the proposed Inception DenseGCN outperforms the feature extractors used in PU-Net and 3PU, when comparing PU-GCN (NodeShuffle) with PU-Net (NodeShuffle) and 3PU (NodeShuffle).
A qualitative comparison of the original 3PU, 3PU (NodeShuffle), and PU-GCN is shown in \figLabel \ref{fig:intro-surface_3pu_up}. NodeShuffle shows a clear improvement in generating samples with less noise and better details, as compared to the original upsampling method used in 3PU. Inception DenseGCN further improves the performance by preserving better intricate structures. 

\begin{table}[ht]
\centering
\caption{\textbf{Ablation study on NodeShuffle.}
Results show that NodeShuffle can transfer well to different upsampling architectures in the literature. Replacing the original upsampling module with NodeShuffle improves upsampling performance overall. The effectiveness of the GCN layer in NodeShuffle is shown when replacing the GCN layer with a set of MLPs (MLPShuffle).  
\vspace{-10pt}
}
\resizebox{\columnwidth}{!}{%
\begin{tabular}{l|ccccc}
\toprule
\multirow{2}{*}{\textbf{Network}}
& \multicolumn{1}{c}{\textbf{CD}} & \multicolumn{1}{c}{\textbf{HD}} & \multicolumn{1}{c}{\textbf{P2F}}&
\multicolumn{1}{c}{\textbf{Param.}} &
\multicolumn{1}{c}{\textbf{Time}}
\\
& \multicolumn{1}{c}{$10^{-3}$} & \multicolumn{1}{c}{$10^{-3}$} & \multicolumn{1}{c}{$10^{-3}$} &
\multicolumn{1}{c}{Kb} &
\multicolumn{1}{c}{ms}
\\
\midrule
PU-Net (Original) \cite{yu2018pu} & 1.155& 15.170& 4.834 & 814.3 & 10.04\\
PU-Net (NodeShuffle)                               &0.974 & 13.522 & 4.474 & 462.1 & 11.04\\
\midrule
3PU (Original) \cite{Yifan20193pu}    & 0.935 & 13.327 & 3.551 & 76.2 & 10.86\\
3PU (NodeShuffle) & 0.780 & 10.462 & 3.228 & \textbf{71.4} & 11.78\\
\midrule
PU-GCN (Duplicate) & 0.788 & 11.269 & 3.031 & 74.2  & \textbf{8.63}\\
PU-GCN (MLPShuffle) & 0.682 & 10.692 & 2.586 & 76.2 & 8.87 \\
PU-GCN (NodeShuffle) & \textbf{0.585}       & \textbf{7.577}       & \textbf{2.499}       & 76.0 & 8.83 \\ 
\bottomrule
\end{tabular}
}
\label{tab:up_ablation_results}
\vspace{-10pt}
\end{table}

\subsection{Effects of Additive Noise and Input Sizes}
\begin{figure}[htb!]
\centering
\includegraphics[page=1,trim = 60mm 0mm 63mm 0mm, clip, width=0.8\columnwidth]{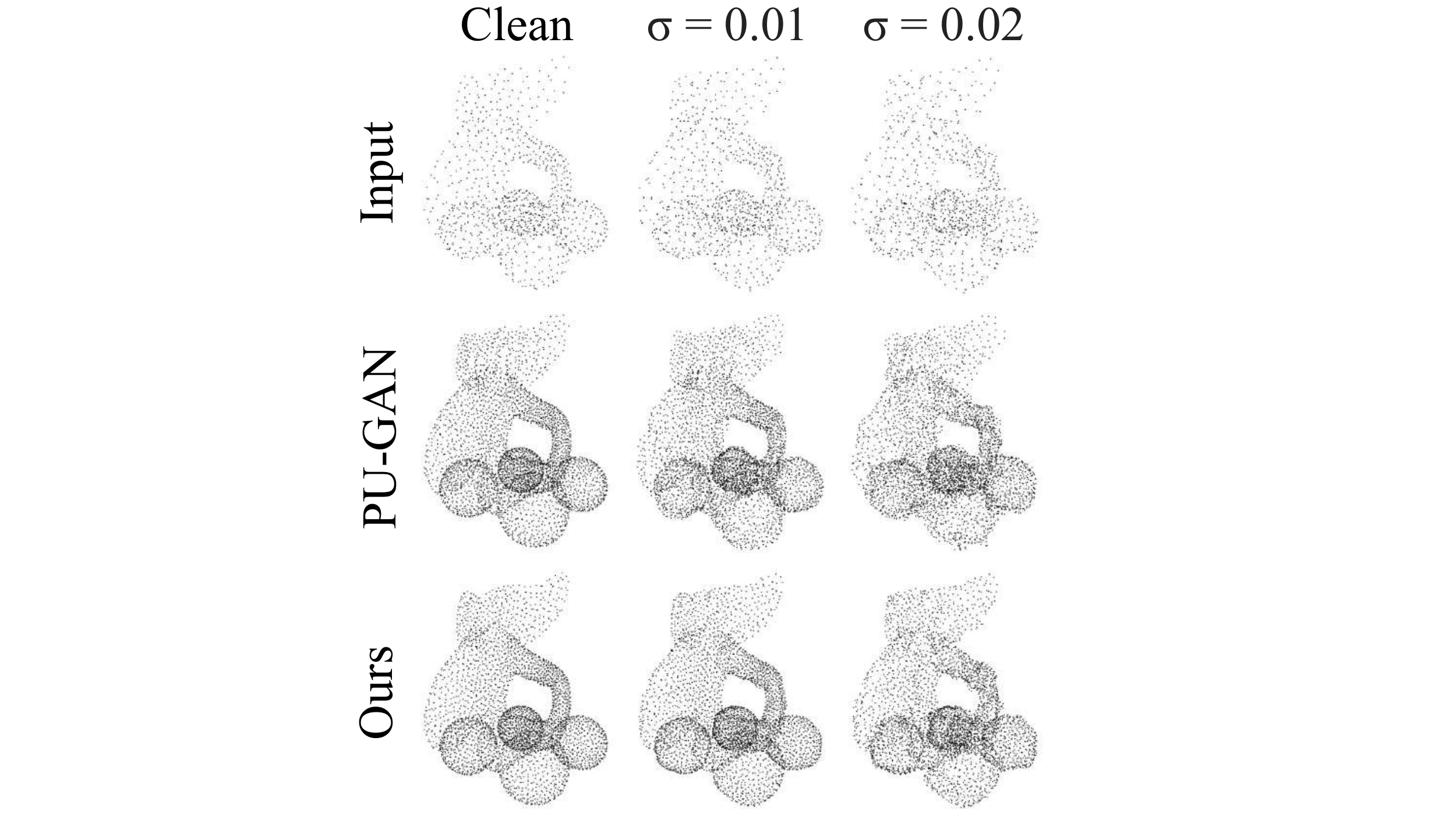}
\caption{\textbf{Effect of additive noise.} The top row shows the input point clouds with different additive noise levels. The middle and bottom rows show the point clouds upsampled by PU-GAN\cite{li2019pugan} and our PU-GCN, respectively. PU-GCN preserves more fine-grained details with fewer outliers. 
}
\label{fig:ablation-upsampling_noise}
\vspace{-10pt}
\end{figure}

\vspace{-5pt}
\mysection{Upsampling noisy point clouds}
To show the robustness of PU-GCN, we perturb the input point cloud with additive Gaussian noise at varying noise levels. We only compare PU-GCN against PU-GAN\cite{li2019pugan} since PU-GAN is the state-of-the-art. Both models are trained using the same augmentation strategy of point cloud perturbation. Qualitative results in \figLabel \ref{fig:ablation-upsampling_noise} show that PU-GCN can preserve fine-grained details with very few outliers, even in the presence of additive noise, while PU-GAN tends to produce more outliers. 

\begin{figure}[htb!]
\centering
\includegraphics[page=2,trim = 45mm 0mm 63mm 0mm, clip, width=0.9\columnwidth]{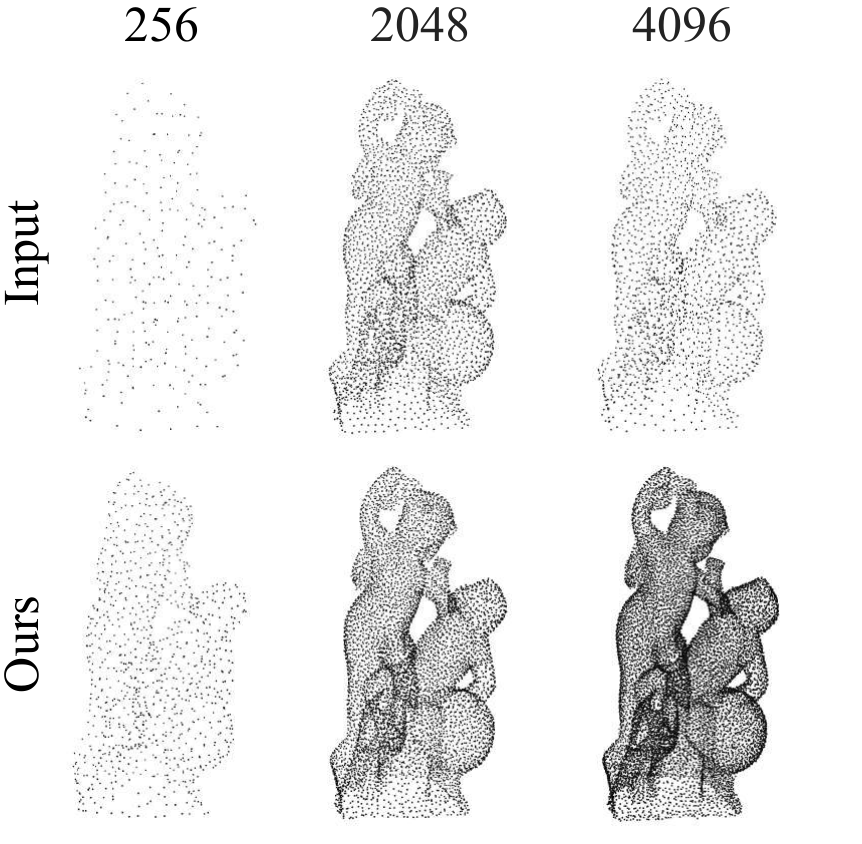}
\caption{\textbf{Effect of input size.} The top row shows the input point clouds of different sizes, and the bottom row shows the $\times4$ upsampled outputs of our PU-GCN. PU-GCN always produces high quality results regardless of input size.  
}
\label{fig:ablation-upsampling_different_size}
\vspace{-10pt}
\end{figure}

\mysection{Upsampling point clouds of varying sizes}
\figLabel \ref{fig:ablation-upsampling_different_size} shows  qualitative examples of upsampling point clouds with PU-GCN for different input sizes. Our PU-GCN always produces high quality point clouds for the range of input sizes. 
This indicates that even if PU-GCN is trained on patches with only 256 points, it can generalize to point clouds with different sizes. 
As expected, PU-GCN generates better quality results when the input point cloud is denser.

\section{Conclusion}
We propose a novel GCN based point cloud upsampling module called NodeShuffle, which improve state-of-the-art upsampling pipelines when it is used in place of the original upsampling. We also introduce the Inception DenseGCN to encode multi-scale information. We further compile and introduce a new large-scale dataset PU1K for point cloud upsampling. Extensive experiments demonstrate that our proposed PU-GCN pipeline, which integrates Inception DenseGCN and NodeShuffle, outperforms state-of-the-art methods on PU1K and another dataset, while requiring fewer parameters and being more efficient in inference. We also show that PU-GCN produces higher upsamping quality on real-scanned point clouds compared to other methods.

\mysection{Acknowledgments}
The authors thank Silvio Giancola and Chen Zhao for their suggestive comments. This work was supported by KAUST Office of Sponsored Research through the Visual Computing Center (VCC) funding. 

{\small
\bibliographystyle{ieee_fullname}
\bibliography{egbib}
}

\end{document}